\def\year{2022}\relax
\documentclass[letterpaper]{article} 
\usepackage{aaai22}  
\usepackage{times}  
\usepackage{helvet}  
\usepackage{courier}  
\usepackage[hyphens]{url}  
\usepackage{graphicx} 
\usepackage{natbib}  
\usepackage{caption} 
\DeclareCaptionStyle{ruled}{labelfont=normalfont,labelsep=colon,strut=off} 
\usepackage{algorithm}
\usepackage{algpseudocode}

\usepackage{newfloat}
\usepackage{listings}
\floatstyle{ruled}
\newfloat{listing}{tb}{lst}{}
\floatname{listing}{Listing}
\usepackage{amsmath}
\usepackage{amssymb}
\usepackage{bbm}
\usepackage{bm}
\usepackage{xspace}
\usepackage{graphicx}
\usepackage{subcaption}
\usepackage{booktabs}
\usepackage{multirow}
\usepackage{xcolor}
\usepackage{CJK}

\newcommand{\method}{KoK\xspace}
\newcommand{\knnmt}{KNN-MT\xspace}
\newcommand{\knn}{KNN\xspace}
\newcommand{\pretrain}{Pre-Train Model\xspace}

\newcommand{\tokenknn}{Token-KNN\xspace}
\newcommand{\lambdaknn}{Policy-KNN\xspace}

\newcommand{\pnmt}{p_{\text{NMT}}\xspace}
\newcommand{\pknn}{p_{\text{KNN}}\xspace}

\newcommand{\rknn}{\mathcal R}
\newcommand{\kknn}{\bm k_{i}\xspace}
\newcommand{\vknn}{v_{i}\xspace}

\newcommand{\rtok}{\mathcal R^{\text{tok}}\xspace}
\newcommand{\ktok}{\bm k^{\text{tok}}_{i}\xspace}
\newcommand{\vtok}{v^{\text{tok}}_{i}\xspace}

\newcommand{\rl}{\mathcal R^{\lambda}\xspace}
\newcommand{\kl}{\bm k^{\lambda}_{i}\xspace}
\newcommand{\vl}{v^{\lambda}_{i}\xspace}

\newcommand{\x}{\bm x}
\newcommand{\h}{\bm h}
\newcommand{\y}{\bm y}
\newcommand{\softmax}{\operatorname{softmax}}
\newcommand{\argmax}{\operatorname{argmax}}
\newcommand{\Transformer}{\operatorname{Transformer}}

\title{Non-Parametric Online Learning from Human Feedback \\ for Neural Machine Translation}
\author {
    Dongqi Wang\textsuperscript{\rm 1},
    Haoran Wei\textsuperscript{\rm 2},
    Zhirui Zhang\textsuperscript{\rm 2},
    Shujian Huang\textsuperscript{\rm 1}\thanks{Corresponding author.}, 
    Jun Xie\textsuperscript{\rm 2}, 
    Jiajun Chen\textsuperscript{\rm 1}
}
\affiliations {
    \textsuperscript{\rm 1}National Key Laboratory for Novel Software Technology, Nanjing University, China\\
    \textsuperscript{\rm 2}Machine Intelligence Technology Lab, Alibaba DAMO Academy \\ 
    \textsuperscript{\rm 1}wangdq@smail.nju.edu.cn,\;\{huangsj, chenjj\}@nju.edu.cn \\ 
    \textsuperscript{\rm 2}\{funan.whr, zhirui.zzr, qingjing.xj\}@alibaba-inc.com\\
}

\begin{document}

\maketitle

\begin{abstract}

We study the problem of online learning with human feedback in the human-in-the-loop machine translation, in which the human translators revise the machine-generated translations and then the corrected translations are used to improve the neural machine translation (NMT) system. However, previous methods require online model updating or additional translation memory networks to achieve high-quality performance, making them inflexible and inefficient in practice.
In this paper, we propose a novel non-parametric online learning method without changing the model structure.
This approach introduces two $k$-nearest-neighbor (\knn) modules: one module memorizes the human feedback, which is the correct sentences provided by human translators, 
while the other balances the usage of the history human feedback and original NMT models adaptively.  
Experiments conducted on EMEA and JRC-Acquis benchmarks demonstrate that our proposed method obtains substantial improvements on translation accuracy and achieves better adaptation performance with less repeating human correction operations. 

\end{abstract}

\section{Introduction}



\begin{figure}[!htbp]
\centering
\includegraphics[width=0.8\linewidth]{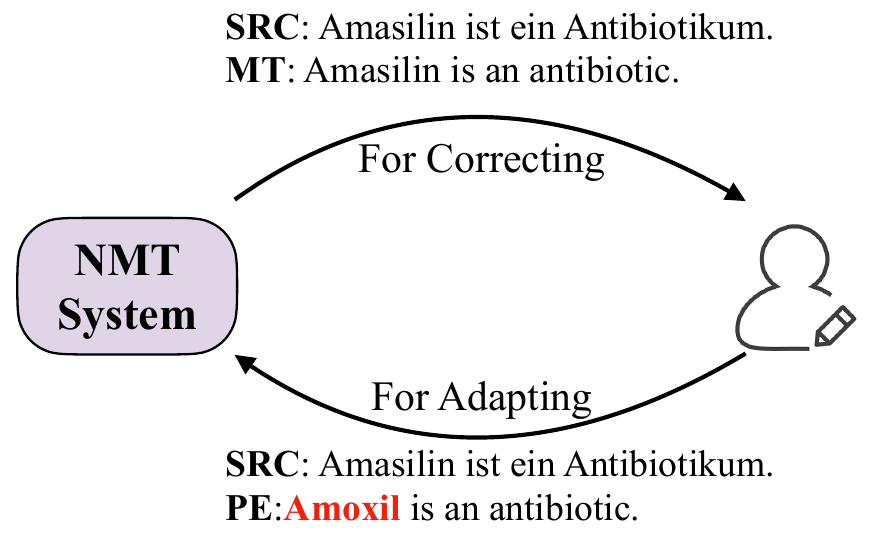}
\vspace{-5pt}
\caption{The workflow of online learning from human feedback in the human-in-the-loop machine translation. Given the source sentence \textit{SRC}, the NMT System generates the translation \textit{MT}. The user corrects it and produces \textit{PE}. Then the corrected sample is used to adapt the NMT system.}
\label{fig:flow}
\end{figure}

The quality of the neural machine translation (NMT) system has been significantly improved recently~\cite{sennrich2015improving,vaswani2017attention,Zhang2018JointTF,Hassan2018AchievingHP}. However, recent research has shown that machine translation still lags behind human parity on translation quality~\cite{Lubli2018HasMT,Freitag2021ExpertsEA}. 
In some scenarios where high-quality translation should be guaranteed,
human-in-the-loop machine translation, i.e., machine translation with human post-editing, is still indispensable. 
As human corrections on machine-translated text are constantly produced, adapting the NMT model to these corrections could improve translation quality and effectively reduce human efforts, as shown in Figure~\ref{fig:flow}.
However, the training of NMT models requires a certain amount of data and consumes dozens or hundreds of GPU hours, which is unavailable in this scenario. Therefore, online learning from human feedback at a low training cost has become a promising research topic in recent years.

A series of studies~\cite{turchi2017continuous,kothur2018document} propose to on-the-fly update the NMT model right after humans correct the translation.
Although these methods improve translation performance, they require extra costs for frequent gradient computations at the inference time. 
Tuning parameters only on the new-coming sentences also brings the risk of catastrophic forgetting problems.
Another research line~\citep{gu2018search,zhang2018guiding} augments the NMT model with the translation memory to cache the human-corrected sentences. 
For each sentence to translate, cached sentences with similar context are retrieved to help the decoding process of the NMT model. 
Memory-augmented NMT avoids updating parameters on-the-fly, but it requires careful network architecture design to model the retrieved translation memories.
Also, extra training overhead is inevitable to optimize the memory network.

Recently, \citet{khandelwal2020nearest} proposed \knnmt, a non-parametric and model-agnostic method for domain adaptation. 
It equips the pre-trained NMT model with a $k$-nearest-neighbor classifier over a datastore of cached context representations and corresponding target tokens.
The final translation probability is the interpolation between the probability calculated by the NMT model and the \knn module. 
\knnmt provides a simple but effective way to exploit human feedback by adding human-corrected translations into the datastore.
However, the success of \knnmt is owed to enough in-domain samples pre-cached in the datastore. 
In the human-in-the-loop translation, to learn the corrected translation, human translators have to make repetitive corrections on the same mistake to provide enough samples for \knnmt.
Biasing the interpolation to the \knn module could accelerate this learning process, but it takes the risk of being interfered by the noise in retrieved neighbors when retrieved items are not relevant enough.
Therefore, it is crucial to balance the usage of the probability from the NMT model and the \knn module when applying \knnmt.

In this paper, we propose KNN-over-KNN (\method), a plug-and-play non-parametric approach for online learning from human feedback. \method equips the NMT model with two KNN modules, \tokenknn and \lambdaknn. \tokenknn is used to model the translation probability as in \knnmt. Human-corrected sentences are incrementally added to the datastore of \tokenknn to improve the translation quality of proceeding sentences. \lambdaknn is introduced to model the balance mechanism between the \tokenknn module and the NMT model. 
It interpolates between the two modules with a Bernoulli distribution. 
The distribution is estimated by a $k$-nearest-neighbor classifier over a datastore.
For building the datastore, we extract features from the  retrieval results of \tokenknn  as the key, and  heuristically induce a binary flag for each result as the value (1 means using this result and vice versa).

We conduct experiments on translating domain-specific documents with various lengths. Oracle references are used to mimic human feedback. 
\method obtains significant improvements (up to 12.9 BLEU improvement and 9.7 TER reduction) compared to the Pre-Trained NMT model.
\method also achieves consistent improvements over existing online learning methods on documents at all lengths. We further show that \method can adapt to human feedback with less repeated corrections, which reduces human effort.
Our code is open-sourced at \url{https://github.com/wangqi1996/KoK}.







\section{Background}

\subsection{Neural Machine translation}
Neural machine translation systems formulate the translation task as a conditional probability model $p(\y|\x)$, which defines the translation process from source sentence $\x=\{x_1, x_2, ..., x_m\}$ into target sentence $\y = \{y_1, y_2, ..., y_{n}\}$. 
The representative works decompose it in an auto-regressive manner from left to right
:
\begin{equation*}
    \pnmt(\y) = \prod_{i=1}^{n} \pnmt(y_t|\y_{<t}, \x)\text{,}
\end{equation*}
where $\y_{<t} = \{y_1, y_2, ..., y_{t-1}\}$ denotes the prefix tokens. Then the probability of each target token is defined as: 
\begin{equation}
\label{eqn:pnmt}
\begin{split}
    \pnmt(y_t|\y_{<t}, \x) &= \softmax(\bm W \bm h_{t} + \bm b) \\
    \bm h_t &= \Transformer(\x, \y_{<t})\text{,}
\end{split}
\end{equation}
where $\bm h_t$ is the representation for context $(\x, \y_{<t})$, and $\bm W$, $\bm b$ are the trainable parameters to mapping the dimension of $\bm h_t$ to the vocab size.

\subsection{Online Learning from Human Feedback}
In the human-in-the-loop translation, a complete translation step is: the NMT model generates machine-translated text, and then the human translator makes revisions on it (human feedback). 
Online Learning from human feedback follows such a paradigm: after the human completes the revision of the current sentence, the machine translation system is adapted incrementally by taking this sample into account.
Concretely, given a document~$\mathcal{D}=\{\bm x^1, \bm x^2, ..,\bm x^{|\mathcal{\bm D}|}\}$, where $\bm x^i$ represents $i^{th}$ sentence, the translation for $\bm x^i$ is generated as
\begin{equation*}
    \hat{\bm y}^i = f^{i-1}(\bm x^{i})
\end{equation*}
where $f^{i-1}$ is the NMT model adapted by $\bm x^{<i}$ sentences and $f^{0}$ is the pre-trained model. 
The human translator corrects $\hat{\bm y}^i$, and then produces the reference $\bm y^i$. The model $f^{i}$ is acquired by adapting $f^{i-1}$ over the new sample $(\bm x^{i}, \bm y^{i})$:
\begin{equation*}
    f^i \leftarrow  \text{Adaptation}(f^{i-1}, \bm x^{i}, \bm y^{i})\text{.}
\end{equation*}
The adaptation process could be tuning the model parameters, or updating the external translation memory. In the rest of the paper, we omit the superscript $i$ for simplicity.

\subsection{\knnmt}

Recently, \citet{khandelwal2020nearest} proposed \knnmt,  a non-parametric method. It augments the NMT model with a token-level $k$-nearest-neighbor retrieval mechanism, allowing it to directly access the cached examples stored in the datastore during inference. It consists of two steps: datastore construction and inference.

\paragraph{Datastore Construction}
The datastore consists of a set of key-value pairs. More specifically, given the parallel sentence in the dataset $(\x, \y) \in (\mathcal{X}, \mathcal{Y})$, the pre-trained NMT model generates the context representation $\bm h_t$ as Equation~\ref{eqn:pnmt} at every timestep $t$, 
then the datastore takes the $\bm h_t$ as the key and the corresponding target tokens $y_t$ as the value:
\begin{equation*}
    (\mathcal{K}, \mathcal{V}) = \{(\bm h_t, y_t), \forall{y_t} \in \y | (\x, \y) \in (\mathcal{X}, \mathcal{Y})\}\text{.}
\end{equation*}

\paragraph{Inference}
The inference process performs in a token-by-token manner. 
Given the source sentence $\x$ and the generated target prefix $\hat{\bm y}_{<t}$, the pre-trained NMT model generates the representation $\hat{\bm h}_t$ and then predicts a probability $\pnmt(\hat y_t|\bm x, \hat{\bm y}_{<t})$ for the target token $\hat y_t$.

Then, the $k$-nearest-neighbor model retrieves the similar $K$ neighbours $\rknn=\{(\kknn, \vknn), i \in \{1,2,...,K\}\}$ in the datastore according to Euclidean distance to $\hat{\bm h}_t$.
The retrieved result is converted into a distribution over the vocabulary by
\begin{equation}
\label{eq:p_knn}
\begin{split}
    p(\kknn|\hat{\bm h}_t) &= \softmax(\frac{-d{(\kknn, \hat{\bm h}_t)}}{T}) \\
    \pknn(\hat y_t| \bm x, \hat \y_{<t}) &= \sum_{(\kknn, \vknn) \in \rknn}\mathbbm{1}_{\vknn=\hat y_t}p(\kknn|\hat{\bm h}_t))\text{,}
\end{split}
\end{equation}
where $d(\bm k_{i}, \hat{\bm h}_t)$ represents the Euclidean distance between $\bm k_{i}$ and $\hat{\bm h}_t$, $p(\bm k_i|\hat{\bm h}_t)$ is the probability of $i^{th}$ retrieval neighbor $(\bm k_{i}, v_{i})$. 
T is a hyper-parameter to control the sharpness of the softmax function. 

The final output distribution is an interpolation between distributions from the NMT model and the KNN retrieved neighbors with a tuned parameter $\lambda \in [0, 1]$
\begin{equation}
\label{eq:all}
\begin{split}
    p(\hat y_{t} |\bm x, \hat{\bm y}_{<t}) &= \lambda *  \pknn(\hat y_t |\bm x, \hat{\bm y}_{<t})  \\
    &+ (1 - \lambda)* \pnmt(\hat y_t |\bm x, \hat{\bm y}_{<t})\text{.}
\end{split}
\end{equation}

\begin{figure}[t]
\centering
\includegraphics[width=1.0\linewidth]{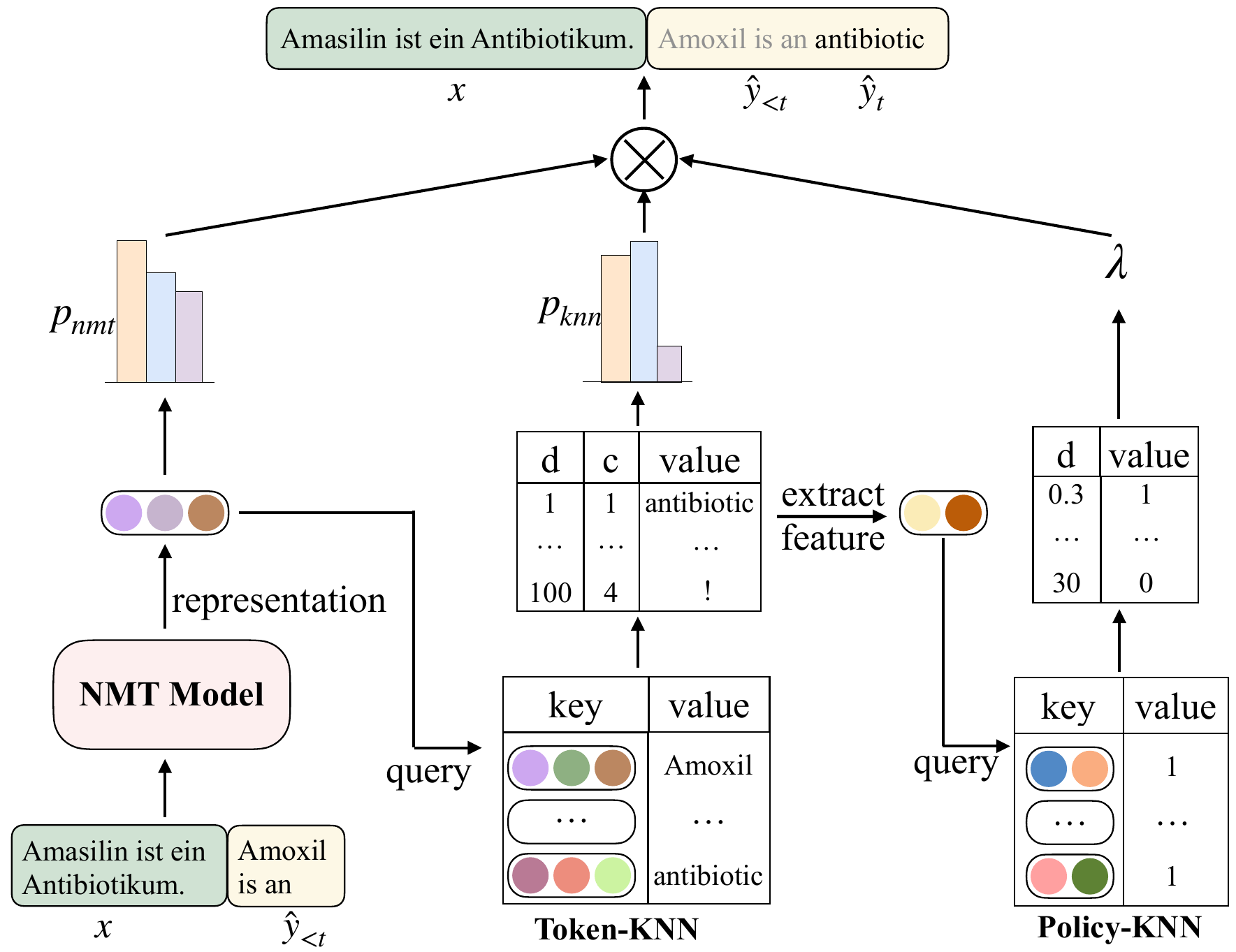}
\caption{An illustration of proposed \method.}
\label{fig:overview}
\end{figure}

\section{Proposed Method}

In this section, we describe the proposed method, KNN-over-KNN (\method). Based on \knnmt, we model a balance mechanism between the usage of the pre-trained NMT model and the \knn module cached history human corrections, i.e., 
we replace the fixed $\lambda$ in Equation \ref{eq:all} into a dynamic $\lambda_t$ varies at each predicting $\hat y_t$:
\begin{equation}
\label{eq:kokall}
\begin{split}
    p(\hat y_{t} |\bm x, \hat{\bm y}_{<t}) &= \lambda_t *  \pknn(\hat y_t |\bm x, \hat{\bm y}_{<t})  \\
    &+ (1 - \lambda_t)* \pnmt(\hat y_t |\bm x, \hat{\bm y}_{<t})\text{.}
\end{split}
\end{equation}
\method learns to predict $\pknn$ and $\lambda_t$ from history human feedback in a non-parametric way, and can be applied to NMT models with various architectures.

\subsection{Architecture}

Figure~\ref{fig:overview} shows the overall architecture of \method. It parameterizes the final translation probability (Equation \ref{eq:kokall}) with two \knn modules besides the pre-trained NMT model, in which the \tokenknn module predicts the probability $\pknn$ over target words  and the \lambdaknn module predicts an interpolation coefficient $\lambda_t$.

During the translation process of one sentence, for each decoding step $t$, \tokenknn takes the hidden representation $\bm {\hat h}_t$ as query and retrieves from its datastore to generate the translation probability $\pknn$. Then features extracted from the retrieval result of \tokenknn are used as the query of \lambdaknn to calculate the $\lambda_t$ value. The final translation probability (Equation~\ref{eq:kokall}) is computed as an interpolation between $\pknn$ and $\pnmt$ with $\lambda_t$. 

\subsection{\tokenknn}

\tokenknn module predicts the translation probability $\pknn$ with $k$-nearest-neighbor classifier over the datastore. 
It is similar to the \knn module in the  \knnmt, but the datastore is built  incrementally. 
During the translation process, with the $\bm{\hat h}_t$ as the query, the \tokenknn queries from the token datastore. The translation probability is computed as Equation~\ref{eq:p_knn}. 
During the adaptation process, given source sentence $\x$ and its human-corrected translation $\y$, we build the key-values pairs and then store them into the datastore.

\subsection{\lambdaknn}

\lambdaknn module calculates the $\lambda_t$, which expresses the balance mechanism  between  the usage of the pre-trained NMT model and the \tokenknn module. 
We introduce a random variable $\bm \lambda$ which obeys the Bernoulli distribution. $\bm \lambda = 1$ represents to use the translation probability of \tokenknn and vice versa for the NMT model. 
Thus, predicting coefficient $\lambda_t$ in Equation~\ref{eq:kokall}
is demonstrated as estimating the conditional probability $p(\bm \lambda  = 1 | \x, \y_{<t})$ 
\begin{equation*}
\begin{split}
     \lambda_t = p(\bm \lambda  = 1 | \x, \y_{<t})\text{.}
\end{split}
\end{equation*}
\lambdaknn estimates this probability with $k$-nearest-neighbor classifier over the datastore as well. In the following, we will describe in detail the datastore construction and inference process of \lambdaknn. At the $t^{th}$ decoding step, we denote the retrieval results from \tokenknn as $\rtok_t =\{(\ktok, \vtok), i \in \{1,2,...,K\}\}$.

\paragraph{Key} 
Refer to ~\citet{Zheng2021AdaptiveNN}, we construct the key vector $\bm s_t$ using two kinds of features extracted from the retrieval results of \tokenknn. One feature is the distance, we denote the Euclidean distance between the context representation $\bm h_t$ and $i^{th}$ retrieval result $\ktok$ as $d_{i}$. 
\begin{equation*}
    d_i = \left\| \bm h_t - \ktok \right\|_2\text{.}
\end{equation*}
The other feature is the counts of distinct values for all retrieval results, marked as $c_{i}$. Specially, $c_{i}$ is computed as:
\begin{equation*}
    c_i = |\text{UNIQUE}(v^{\text{tok}}_1,v^{\text{tok}}_2,...,\vtok)|
\end{equation*}
where $\text{UNIQUE}$ represents the unique elements in the list.
$\bm s_t$ is constructed by concatenating all the features:
\begin{equation}
    \label{eq:lambda_knn_key}
    \bm s_t = [d_{1}, d_{2}, ..., d_{K}; c_{1}, c_{2},..., c_{K}]\text{.}
\end{equation}

Intuitively speaking, the distance features $d_{i}$ stand for similarity, which directly evaluates the importance of each retrieval result. The count features $c_{i}$ represents the consistency of retrieval results, as more distinct retrieved values mean less credible \knn retrieval. Both of these two features are crucial for predicting an optimal value of $\bm \lambda$.

We further re-weight the key vector according to the significance of each element. We adopt exponential re-weighting on distance features and count features, respectively: 
\begin{equation}
    \label{eq:lambda_knn_key_recale}
    \bm s_t = \bm s_t \odot [\frac{1}{4}, \frac{1}{8}, ...,\frac{1}{2^{K}}, \frac{1}{2^{K}}; \frac{1}{2^{K}}, \frac{1}{2^{K}},..., \frac{1}{8}, \frac{1}{4}]\text{.}
\end{equation}

\paragraph{Value}
The retrieval result of \tokenknn does not explicitly come with a golden value of $\bm \lambda$ indicating whether to use this result or not. Instead, we implicitly induce the value of $\bm \lambda$ through a simple heuristic:
\begin{equation}
        \label{eq:lambda_knn_value}
        \bm \lambda  = \begin{cases}
                1 & \pknn(y_t |\x, \y_{<t}) > \pnmt(y_t |\x, \y_{<t})  \\
                0 & \pknn(y_t |\x, \y_{<t}) \le \pnmt(y_t |\x, \y_{<t}) \\
            \end{cases}
\end{equation}
where $\x$ and $\y$ is the source sentence and its human-corrected translation, respectively.

This heuristic is actually mathematically meaningful. To find the value of $\lambda_t$ to maximize the probability of \knnmt (Equation \ref{eq:kokall}), we need to solve the optimization problem below:
\begin{equation*}
\begin{split}
        \lambda_t &= \argmax~p(y_t) \\
                &= \argmax_{\lambda_t}~\lambda_t \cdot  \pknn(y_t) + (1 - \lambda_t) \cdot \pnmt(y_t)  \\
                &= \argmax_{\lambda_t}~\pnmt(y_t) + (\pknn(y_t) - \pnmt(y_t)) \cdot \lambda_t  \text{.}
\end{split}
\end{equation*}

\paragraph{Inference} 
During the translation process, at the $t^{th}$ decoding step, given the retrieval result $\rtok_t$ of \tokenknn,  we build the query feature $\bm {\hat s}_t$ as in Equation~\ref{eq:lambda_knn_key_recale}, then \lambdaknn retrieves $K$ neighborhoods, $\rl_t=\{(\kl, \vl), i \in \{1,2,...,K\}\}$, from its datastore.
The retrieval result $\lambda_t$ is calculated by:
\begin{equation}
\label{eq:lambdaknn_cal}
\begin{split}
    p(\kl| \bm{\hat s}_t) &= \softmax(\frac{-d{(\kl, \bm{\hat s}_t})}{T}) \\
    \lambda_t &= p(\bm \lambda  = 1 | \x, \hat \y_{<t})  \\
    &= \sum_{(\kl, \vl) \in \rl } \mathbb I_{\vl=1} * p(\kl| \bm{\hat s}_t)\text{.}
\end{split}
\end{equation}
The final translation probability is calculated as Equation~\ref{eq:kokall}.

\subsection{Implementation}
Overall, \method is performed as an online learning paradigm. It takes three steps when translating a source sentence $\bm x$:
\begin{enumerate}
    \item[a.] \textbf{Translation}: \method generates the translation $ \hat{\y}$ by maximize Equation~\ref{eq:kokall}. At each decoding step, \method computes $\pnmt$ from the NMT model, $\pknn$ from the \tokenknn, and $\lambda_t$ from the \lambdaknn, respectively.
    
    \item[b.]  \textbf{Correction}: The human translator corrects the translation result $\hat{\y}$ and produces the corrected version $\y$.
    
    \item[c.]  \textbf{Adaptation}: $\bm x$ and $\bm y$ are encoded by the NMT model with teacher-force decoding. All the tokens in $\y$ are used to retrieve from \tokenknn as well. Then we update \lambdaknn and \tokenknn by building key-value pairs, successively.
\end{enumerate}
When the translation of a document is completed, one can choose to clear the datastores of two \knn modules for translating new documents, or keep them for proceeding translation on similar documents, which is dependent on user's demand. We clear the datastores for translating new documents in this paper. The details of the creating/updating process of datastore are shown in Appendix.

\section{Experiments}

\subsection{Experimental Setup}
We evaluate the proposed method by equipping the general-domain pre-trained NMT model with \method and use it to translated in-domain documents with various document sizes.
We use the oracle reference as the human-corrected translations to simulate the real human-in-the-loop scenario.

\paragraph{Dataset.}

We conduct the experiments on two specific domain datasets from OPUS~\cite{Tiedemann2012ParallelDT},  which are widely employed by previous works~\cite{Zheng2021AdaptiveNN, cai2021neural}: (1) European Medicines Agency (EMEA) dataset\footnote{http://opus.lingfil.uu.se/EMEA.php}~\citep{tiedemann2009news}, which consists of sentence-aligned documents focusing on medical products.
(2) JRC-Acquis corpus~\cite{steinberger-etal-2006-jrc}, which contains the European Union laws applicable to the EU member states.
Following common practices, we use the Moses toolkit\footnote{https://github.com/moses-smt/mosesdecoder} to tokenize the documents, remove the same sentences in a document and then segment the words into subword units~\cite{sennrich-etal-2016-neural} with the bpe-codes provided by the pre-trained model.  

We validate our method on documents with various lengths. 
Specifically, we divide the documents into five buckets based on their length (0-50, 50-100, 100-200, 200-500 and 500-1000). We randomly select some documents so that the total document length of each bucket is upper than 1000.
Detailed for EMEA/JRC dataset statistics are shown in Table~\ref{tab:emea-data}.

\begin{table}[tbp]
\centering
\small
\tabcolsep 2.8pt
\begin{tabular}{lccccc}
\toprule
\textit{Bucket}      & 0-50   & 50-100   & 100-200   & 200-500   & 500-1000 \\
\midrule
\multicolumn{6}{c}{EMEA} \\
\midrule
\textit{Documents}             & 22    & 14        & 7         & 4         & 5 \\    
\textit{Ave sentences}    & 38.4 & 73.0     & 157.9    & 392.8    & 759.2 \\
\textit{Ave tokens}       & 1174.7     & 1938.9   & 3466.1   & 9334.5   & 22725.6 \\
\midrule
\multicolumn{6}{c}{JRC-Acquis} \\
\midrule
\textit{Documents}            & 22       & 14        & 7         & 4         & 5\\    
\textit{Ave sentences}    & 38.1     & 73.1      & 158.5     & 373.8     & 734.8\\
\textit{Ave tokens}       &  1347.1  & 2466.7    & 5345.4    & 12518.2   & 26409.2 \\
\bottomrule
\end{tabular}
\caption{Data statistics for the EMEA and JRC-Acquis dataset. \textit{Documents} represents the number of documents in the bucket. \textit{Ave sentences/tokens} stand for the average sentences/tokens of the documents in the bucket.}
\label{tab:emea-data}
\end{table}

\paragraph{Implementation Details.}
We apply the $\textsc{Fairseq}$\footnote{https://github.com/pytorch/fairseq}~\cite{ott2019fairseq} toolkit for NMT implementation, 
and Faiss\footnote{https://github.com/facebookresearch/faiss}\cite{JDH17} with \textit{Exact Search for L2} setting for efficient KNN retrieval.
Following the previous experiences~\cite{Zheng2021AdaptiveNN, khandelwal2020nearest}, we employ the WMT19 German-English news translation task winner model~\cite{ng-etal-2019-facebook} as the pre-trained model. The $K$ for \tokenknn and \lambdaknn is 8.

\begin{table*}[t]
\centering
\small
\tabcolsep 5pt
\begin{tabular}{lcccccccccccc}
\toprule
\multirow{2}*{\textit{Bucket}} & \multicolumn{2}{c}{0-50}   & \multicolumn{2}{c}{50-100}   & \multicolumn{2}{c}{100-200}   & \multicolumn{2}{c}{200-500}   & \multicolumn{2}{c}{500-1000} & \multicolumn{2}{c}{Average}\\
                    & BLEU  & TER   & BLEU  & TER   & BLEU  & TER   & BLEU  & TER    & BLEU  & TER  & BLEU & TER \\
\midrule
\textit{Pre-Trained}           & 43.8  & \textbf{52.1} & 43.1  & 52.8  & 38.3  & 54.0  & 41.9  & 53.8 & 40.8  & 53.4  & 41.6 & 53.2  \\
\textit{Online Tuning}       & 44.0 & 52.2 & 43.5 & 52.3 & 39.6 & 51.4 & 43.8 & 51.8 & 44.7 & 49.3 & 43.1 & 51.4   \\
\textit{\knnmt}              & 43.8 & 52.6 & 43.6 & 52.5 & 40.0 & 53.1 & 43.8 & 52.3 & 44.2 & 50.8 & 43.1 & 52.3  \\      
\textit{Adaptive \knnmt}     & 29.7 & 70.2  & 28.9 & 70.3  & 35.9 & 58.4  & 37.2 & 61.2  & 48.2 & 50.3  & 36.0 & 62.1 \\
\midrule
\textit{\method}             & \textbf{44.4} & \textbf{52.1}  & \textbf{43.9} & \textbf{52.4}  & \textbf{44.1} & \textbf{50.0}  & \textbf{45.7} & \textbf{51.1} & \textbf{53.7} & \textbf{43.7} & \textbf{46.4} & \textbf{49.9} \\
\bottomrule
\end{tabular}
\caption{BLEU~($\uparrow$) and TER~($\downarrow$) on the EMEA dataset. }
\label{tab:medical}
\end{table*}

\begin{table*}[t]
\centering
\small
\tabcolsep 5pt
\begin{tabular}{lcccccccccccc}
\toprule\
\multirow{2}*{\textit{Bucket}} & \multicolumn{2}{c}{0-50}   & \multicolumn{2}{c}{50-100}   & \multicolumn{2}{c}{100-200}   & \multicolumn{2}{c}{200-500}   & \multicolumn{2}{c}{500-1000} & \multicolumn{2}{c}{Average}\\
                    & BLEU  & TER       & BLEU  & TER   & BLEU  & TER   & BLEU  & TER    & BLEU  & TER  & BLEU & TER \\
\midrule
\textit{Pre-Trained} & 54.0 & 37.2 & 49.9 & 41.2 & 41.9 & 47.1 & 39.9 & 48.8 & 43.4 & 45.3 & 45.8 & 43.9 \\
\textit{Online Tuning}  & 54.4 &  37.0 & 50.9 &  40.5 & 43.8 &  45.4 & 42.8 &  46.3 & 47.5 &  42.2 & 47.9 & 42.3    \\
\textit{\knnmt}  & 55.5 & 	38.0 & 52.2 & 	40.9 & 45.7 & 	44.7 & 43.6 & 	46.4 & 47.7 & 	42.4 & 48.9 & 	42.5  \\      
\textit{Adaptive \knnmt} & 42.5 & 53.0 & 41.7 & 52.8 & 40.2 & 52.3 & 39.2 & 52.4 & 45.4 & 46.3 & 41.8 & 51.3 	 \\
\midrule
\textit{\method}            & \textbf{56.3} & \textbf{35.1}  & \textbf{52.4} & \textbf{39.3}  & \textbf{47.7} & \textbf{43.1}  & \textbf{44.7} & \textbf{45.6}  & \textbf{50.1} & \textbf{40.1}  & \textbf{50.2} & \textbf{40.6}  \\
\bottomrule
\end{tabular}
\caption{BLEU~($\uparrow$) and TER~($\downarrow$) on the JRC-Acquis dataset. }
\label{tab:law}
\end{table*}

\paragraph{Baselines.}
We compare our method with several representative researches, including:
\begin{itemize}
    \item Pre-Trained: We only use the pre-trained NMT model during translation, which measures the domain diversity without any adaptation.
    \item Online Tuning: Online updating the pre-trained NMT model with human-corrected sentences. Following \citet{kothur2018document}, we train the model 5 steps on every single sentence with the adam optimizer. The hyper-parameters setting follows the pre-trained models,  except the learning rate is set to $1.0\times 10^{-7}$ and fixed during the inference process.
    \item \knnmt: 
    After every sentence is translated, we add the human-corrected sentences to the datastore of \knnmt.
    Similar to our method, the $K$ is 8. The $\lambda$ value for the EMEA dataset and the JRC-Acquis dataset are 0.2 and 0.3, respectively.
    \item Adaptive \knnmt~\cite{Zheng2021AdaptiveNN}: A variant of \knnmt, which introduces a network to dynamically determine the number of $K$ for each target token. We use the model pre-trained on the IT domain~(provided by the paper) and incrementally update the datastore as in the \knnmt.
\end{itemize}

\paragraph{Evaluation.}

We use SacreBLEU\footnote{https://github.com/mjpost/sacrebleu}~\cite{post-2018-call} to measure the result with case-sensitive detokenized BLEU~\citep{papineni-etal-2002-bleu}. We concatenate the translations of all the documents in each bucket and calculate the corpus-level BLEU.
For the completeness of the results, we also report the TER~\cite{snover-etal-2006-study} metric to compute the edit distance between the reference and system translation.

\subsection{Main Results}

We can see that consistent improvements (up to 12.9 BLEU improvement and 9.7 TER reduction) are achieved by \method on different document lengths. As a comparison, Online Tuning achieves relatively  minor improvement on long documents, probably because that several thousand samples are still not enough for a gradient-based approach. 
\knnmt fails to attain similar improvements as \method on long documents because of the small but fixed $\lambda$ value. This indicates the benefit of adjusting $\lambda$ adaptively.
Instead, the performance of adaptive \knnmt deteriorates when the length of a document is relatively small.
As the training of the adaptive \knnmt is based on a static datastore built from all in-domain training data, it is not compatible with the human-in-the-loop machine translation scenario that the human feedback comes in a streaming way.
Overall, the results demonstrate the effectiveness and generalization of our method on different document conditions.

We also conduct experiments on the JRC-Acquis dataset, which contains documents related to the law domain. All the results are listed in Table~\ref{tab:law}. \method consistently outperforms all the baselines, which demonstrates that our approach can be generalized to other domains.
We show the performance on the full test sets in Appendix.

\subsection{Comparing With Translation Memory}

We also compare our method with the NMT model augmented by the translation memory (TM-NMT)~\cite{kuang2020translation, bapna2019non}. Due to the training data of the WMT19 NMT model is unavailable, we implement the TM-NMT model and the NMT model of \method on the WMT14 DE-EN dataset for a fair comparison, and evaluate them on the EMEA dataset. For the NMT model, we use the transformer big architecture, and the training detail follows the introduction of fairseq\footnote{https://github.com/pytorch/fairseq/issues/202}. For the TM-NMT, we follow the \textit{source similarity} method in~\citet{cai2021neural}, and the architecture is the same as the transformer big model.

The result is illustrated in Table~\ref{tab:TM}. As seen, \method still achieves improvements on all the document sizes. However, TM-NMT fails to exceed the pretrained NMT model except on the 200-500 bucket, which may be resulted from the inability to handle the low quality of retrieval memories~\cite{cai2021neural}.

\begin{table*}[tbp]
\centering
\small
\tabcolsep 5pt
\begin{tabular}{lcccccccccccc}
\toprule\
\multirow{2}*{\textit{Bucket}} & \multicolumn{2}{c}{0-50}   & \multicolumn{2}{c}{50-100}   & \multicolumn{2}{c}{100-200}   & \multicolumn{2}{c}{200-500}   & \multicolumn{2}{c}{500-1000} & \multicolumn{2}{c}{Average}\\
                    & BLEU  & TER       & BLEU  & TER   & BLEU  & TER   & BLEU  & TER    & BLEU  & TER  & BLEU & TER \\
\midrule
\textit{Pre-Trained}   & 38.8 & 55.7 & 36.6 & 57.1 & 33.2 & 59.2 & 33.9 & 60.1 & 32.2 & 62.0 & 29.1 & 58.8 \\
\textit{TM-NMT} & 38.1 & \textbf{55.6} & 35.6 & 57.7 & 32.7 & 59.0 & 35.9 & 59.8 & 32.0 & 61.8 & 29.1 & 58.8 \\      
\midrule
\textit{\method}   & \textbf{39.0} & \textbf{55.6} & \textbf{37.3} & \textbf{56.7} & \textbf{36.6} & \textbf{56.6} & \textbf{38.6} & \textbf{57.6} & \textbf{44.1} & \textbf{52.5} & \textbf{32.6} & \textbf{55.8}        \\
\bottomrule
\end{tabular}
\vspace{-3pt}
\caption{Performance comparison between \method and NMT model augmented by the translation memory (TM-NMT). }
\label{tab:TM}
\end{table*}

\begin{table}[!htbp]
\centering
\small
\tabcolsep 5pt
\begin{tabular}{lcc}
\toprule\
\textit{Model}       &  Latency  & Speedup \\
\midrule
\textit{Pre-Trained} &  269.5 ms & $\times 1.00$   \\
\textit{Online Tuning} &  289.9 ms & $\times 0.93$   \\
\textit{\knnmt} &   324.7 ms & $\times 0.83$ \\
\textit{Adaptive \knnmt} &  409.8 ms & $\times 0.66$  \\
\midrule
\textit{\method} &   384.6 ms & $\times 0.70$ \\
\bottomrule
\end{tabular}
\vspace{-3pt}
\caption{Decoding latency of different models on the EMEA dataset. We also list the speedup w.r.t the pre-trained NMT model.}
\label{tab:speed}
\end{table}

\subsection{Latency}

In the human-in-the-loop translation scenario, human translators are sensitive to decoding latency. Therefore, we also evaluate the latency of the proposed \method.
We simulate the Computer Aided Translation(CAT) task setting and measure the decoding latency sentence-by-sentence\footnote{The latency is measured on a single GeForce GTX 1080-Ti GPU.}. The latency is calculated on the EMEA dataset, and we report the average latency in milliseconds.
The speed-up ratio is also computed by comparing it with the \pretrain following the previous practice~\citep{Zheng2021AdaptiveNN}. 
The result is summarized in Table~\ref{tab:speed}. Our method slightly slows down the inference time, but significantly improves translation quality, which is tolerable for users.




\section{Analysis}
In this section, we perform several analyses and ablation studies for the proposed \method.

\subsection{Can \method Learn Optimal $\lambda$}





\method equips \knnmt with a \lambdaknn to adaptively decide between adopting knowledge from human feedback or keeping decoding results from the pre-trained NMT model on every token. In this section, we demonstrate that \method indeed learns an optimal policy to make the decision.

\begin{figure}[h]
\centering
\includegraphics[width=0.82\linewidth]{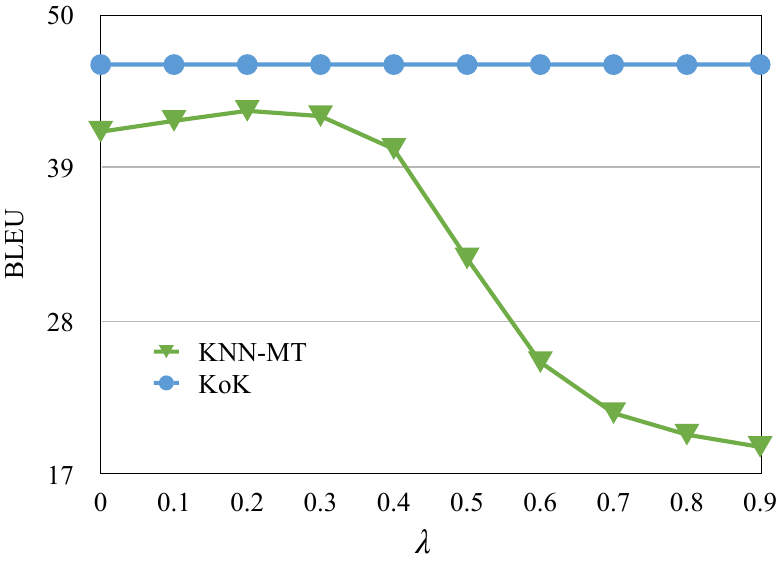}
\vspace{-6pt}
\caption{Results of \method v.s. \knnmt with different $\lambda$ values on EMEA datasets.}
  \label{fig:lambda}
\end{figure}

We first compare \method with \knnmt which chooses different $\lambda$ values. The results are shown in Figure~\ref{fig:lambda}.
We try $\lambda \in [0.0, 1.0)$ with 0.1 as the step size, and report average BLEU scores on the various document length buckets on the EMEA dataset.
\knnmt is sensitive to the choice of $\lambda$ value (BLEU varies from 18 to 43), which indicates the importance of selecting  a proper $\lambda$ value.
\knnmt achieves the best performance when $\lambda$ is around $0.2$, but still lags behind \method. 
This experiment shows that it is crucial to decide the $\lambda$ value token-by-token adaptively.



\begin{figure}[h]
\centering
\includegraphics[width=0.82\linewidth]{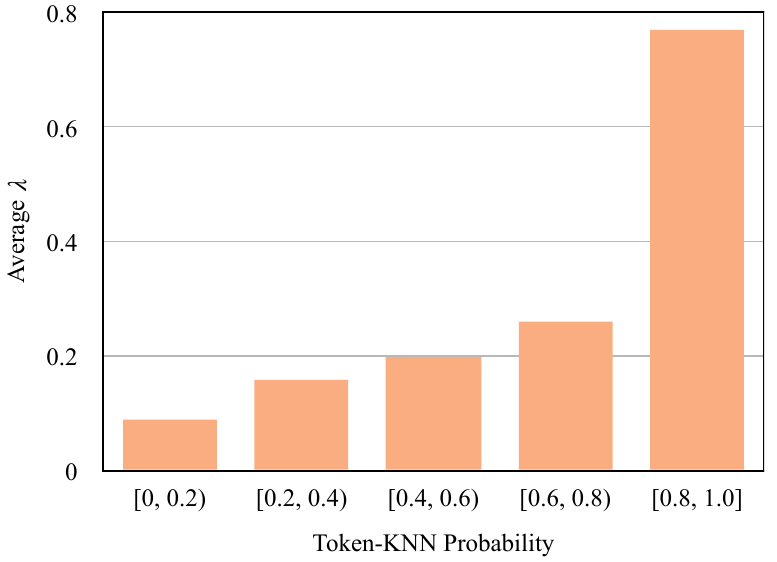}
\vspace{-6pt}
\caption{Average $\lambda$ in different ranges of reference \tokenknn probability. We compute the \tokenknn probability and its corresponding $\lambda$ value by \lambdaknn for each words in the reference. }
 \label{fig:similar}
\end{figure}

We further evaluate whether \method could learn an optimal $\lambda$. Given oracle target $\y$, for each $y_t \in \y$, we compute the \tokenknn probability of $y_t$ as $p_{knn}(\y_t)$, and the corresponding $\lambda_t$ predicted by \lambdaknn. To show the relationship between $\lambda_t$ and $p_{knn}(\y_t)$, we divide all the \tokenknn probability values into five buckets and compute the average $\lambda$ values in every bucket. The result is shown in Figure~\ref{fig:similar}. As seen, the $\lambda$ value becomes high when the \tokenknn probability gets bigger, and vice versa. This is in line with the motivation that \method should bias to the usage of  \tokenknn probability when retrieved human feedbacks indeed help the translation.
We also evaluate the effectiveness of K in the Appendix.

\subsection{Zero-Shot and Few-Shot Ability of \method}
In machine translation with the human-in-the-loop scenario, users not only care about the quality of the translation model, but also care about how fast the model can learn from their feedback to avoid repeatedly making the same mistakes they already corrected before. Therefore, in this section, we evaluate the adaptation speed of \method to user feedback.

We follow the R-indicator used in ~\citet{measuring}, which measures the translation recall of words with different occurrence times in users' feedback. We denote $R_i$ as the recall of tokens that have appeared $i$ times in the previous corrected sentences:
\begin{equation}
\label{eq:Ri}
    R_i = \frac{\sum_{j=1}^{|\mathcal{D}|} |\mathcal{H}_j \cap \mathcal{R}_{i,j}|}{\sum_{j=1}^{|\mathcal{D}|} |\mathcal{R}_{i,j}|}
\end{equation}
where $\mathcal{H}_j$ represents unique words in the $j^{th}$ machine-generated translation and $\mathcal{R}_{i,j}$ represents unique words in the $j^{th}$ reference that are their $(i + 1)^{th}$ occurrence in the whole document. Specifically, $R_0$ evaluates the tokens that first appear in the translating document and $R_1$ considers those that have appeared once.

\begin{figure}[t]
\centering
\includegraphics[width=0.85\linewidth]{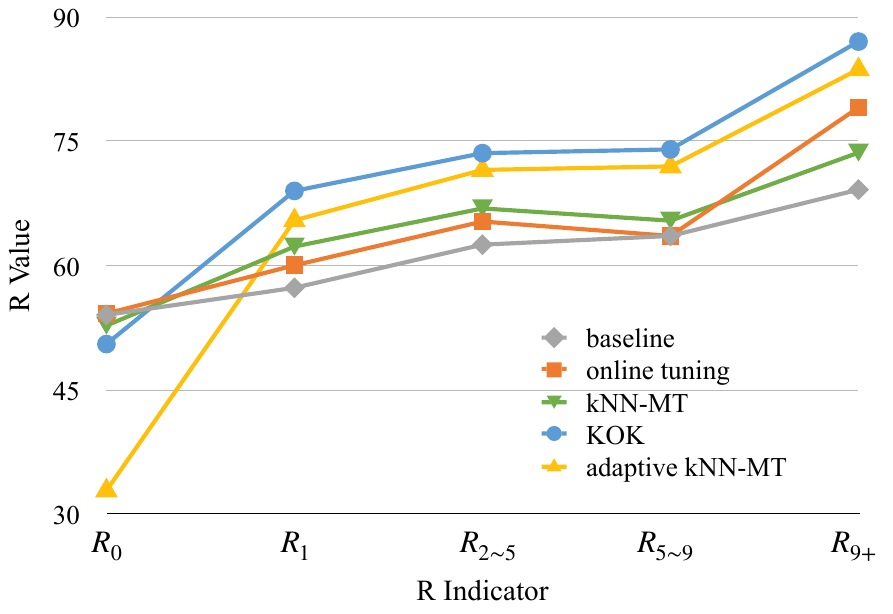}
\vspace{-6pt}
\caption{Results on documents with $[200,500]$ bucket from EMEA dataset for the proposed R-indicator.}
\label{fig:R}
\end{figure}

We conduct experiments on documents with $[200,500]$ bucket from EMEA. $R_0$, $R_1$, $R_{2\sim5}$, $R_{5\sim9}$ and $R_{9+}$ are computed for \method and other baselines as Equation \ref{eq:Ri} respectively ($R_{m\sim n}$ is defined as the micro average of $R_m, R_{m+1},...,R_n$). The results are shown in Figure~\ref{fig:R}. \method achieves the comparable $R_0$, but the $R_i$ values improve quickly and outperform all the other methods.
It indicates \method's ability to reduce interference from unrelated retrieval results of \tokenknn, and to adapt to helpful human feedback faster. Adaptive \knnmt achieves a lower $R_0$, which may due to the same reason about worse performance in main experiments.


\section{Related Work}

\paragraph{Online Learning from Human Feedback. }

Online updating~\cite{peris2019online, turchi2017continuous, wei2019correct} the NMT model has been proved to be an effective and efficient method to adapt the model to the target domain. 
\citet{kothur2018document} proposed the single-sentence adaptation, which performed online adaptation one sentence at a time. Simultaneously, they introduced the dictionary adaptation to address the problem of translating novel words.
\citet{karimova2018userstudy, domingo2019incremental, domingo-etal-2020-user} presented a user study to prove the advantages of online adaptation.
\citet{domingo2019demonstration} applied online adaptation technology in a production environment.
\citet{mueller2019sentence} proposed to fine-tune the model over a subset of similar sentences extracted from the training set. 

However, these methods require updating the model frequently, which brings a non-negligible computation cost in the human-in-the-loop translation scenario, and is prone to catastrophic forgetting problems.

\paragraph{Translation Memory.}
The NMT model augmented with the translation memory \cite{kuang2020translation, bapna2019non} followed the generate-then-retrieve manner.
\citet{tu2018learning} proposed to retrieve samples similar to the source sentence and then encoded retrieved source-target pairs using key-value memory networks.
\citet{cai2021neural} proposed to store the target-language sentence in the memory and used a cross-lingual retrieval model.
\citet{cao-xiong-2018-encoding, cao2019learning} designed gating mechanisms to balance the impact of the translation memory.
\citet{zhang2018guiding} up-weighted the NMT output with the retrieved n-gram.
\citet{bulte-tezcan-2019-neural, xu-etal-2020-boosting} augmented NMT training data with fuzzy TM matches.
\citet{Xia_Huang_Liu_Shi_2019} packed the TM into a graph and using the attention mechanisms to integrating the TM representation into the NMT model.
However, these methods need to modify the architecture carefully to leverage the retrieved similar sentences, for example, adding extra components into the decoder of the NMT model to encoding and incorporate the retrieved sentences.

A series of research incorporated the knowledge into NMT systems through a non-parametric method.
\citet{khandelwal2019generalization} proposed \knn-LM, augmented language model with the retrieved similar sentences using a \knn model.
\citet{khandelwal2020nearest} generated the translation with the nearest neighbor classifier over a large datastore of cached examples.
\citet{Zheng2021AdaptiveNN} proposed to dynamically determine the number of $K$ for each target token.
\citet{Zheng2021NonParametricUD} further extended the capability of \knn-MT on unsupervised domain adaptation setting.

\section{Conclusion}
In this paper, we propose KNN-over-KNN (\method), a plug-and-play non-parametric method for online learning from human feedback. \method introduces two KNN modules, one to memorize the corrected sentences by the human and the other to balance the importance of the in-domain corrected sentences and the general-domain pretrain NMT model.
In the experiments, our method significantly improves the performance and achieves better performance than state-of-the-art baselines on one-shot or few-shot learning for domain-specific lexical items.
As \method achieves competitive performance compared with existing online learning methods, applying our approach to other generation tasks is a promising direction for future work.


\section*{Acknowledgements}
We would like to thank the anonymous reviewers for their insightful comments. This work is supported by National Science Foundation of China (No. U1836221, 6217020152) and National Key R\&D Program  of  China  (2018YFB1403202). The work was done when the first author was an intern at Alibaba Group.

\bibliography{aaai22}

\appendix

\clearpage
\setcounter{page}{1}
\setcounter{table}{0}
\setcounter{figure}{0}
\setcounter{footnote}{0} 

\section{The Creating/Updating Process of Datastore}\label{append:datastore}
At the beginning of the inference process, the datastore is empty. For each source sentence $\x$ and the human corrected sentence $\y$, the adaptation process is:
\begin{itemize}
    \item The pre-trained model generates the hidden representation $\h$ by teacher forcing.
    \item With $\h$ as the query, the retrieval result of \tokenknn is $\rtok$.
    \item The items for the datastore of \lambdaknn are extracted from $\rtok$ as described in the subsection \textit{\lambdaknn} in page 3.
    \item The items for the datastore of \tokenknn are $\h$ and the corresponding target token $\y$.
    \item These items are added into the datastores of \tokenknn and \lambdaknn respectively.
\end{itemize}
The \knn models are non-parametric and don't need pre-training. We will make it clearer in the next version.

\section{The Performance on The Full Test Sets}
We calculate the average BLEU score over the buckets in the main paper, and re-calculate the BLEU scores on the full test sets. The scores are listed in Table~\ref{tab:all_test}. We will add these results in the next version.

\begin{table}[htbp]
\centering
\begin{tabular}{lcc}
\toprule
\textit{Dataset} & EMEA & JRC-Acquis \\ 
\midrule
\textit{Pre-trained} & 41.6 & 44.5 \\
\textit{Online Tuning} & 43.8 & 47.3 \\
\textit{\knnmt} & 43.5 & 48.1 \\
\textit{Adaptive \knnmt} & 40.5 & 42.9 \\
\textit{\method} & 49.1 & 49.7 \\
\bottomrule
\end{tabular}
\caption{BLEU~($\uparrow$) calculated on full test sets.}
\label{tab:all_test}
\end{table}

\section{The Effectiveness of K}
We try different $K$ values both on \knnmt and our method on documents with [100,200] bucket from EMEA dataset. The results are listed in Table~\ref{tab:K}. The results show that our method is robust to the selection of $K$ values.  We will add these results in the next version.

\begin{table}[htbp]
\centering
\begin{tabular}{lcc}
\toprule
$K$ & \knnmt & \method \\ 
\midrule
1 & 39.3 & 40.1 \\
2 & 39.4 & 42.2 \\
4 & 39.8 & 43.4 \\
8 & 40.0 & 44.1 \\
16 & 39.9 & 44.3 \\ 
32 & 39.8 & 44.2 \\ 
\bottomrule
\end{tabular}
\caption{BLEU~($\uparrow$) of \knnmt and \method with different $K$ values on documents with [100,200] bucket from EMEA dataset.}
\label{tab:K}
\end{table}
\end{document}